\documentclass{article}
\usepackage{ijcai19}

\usepackage{times}
\usepackage{soul}
\usepackage{url}
\usepackage[hidelinks]{hyperref}
\usepackage[utf8]{inputenc}
\usepackage[small]{caption}
\usepackage{graphicx}
\usepackage{amsmath,bm}
\usepackage{booktabs}
\usepackage{graphics} % for pdf, bitmapped graphics files
\usepackage{epsfig} % for postscript graphics files
\usepackage{mathptmx} % assumes new font selection scheme installed
\usepackage{times} % assumes new font selection scheme installed
\usepackage{amssymb}  % assumes amsmath package installed

\usepackage{multicol}
\usepackage{multirow}
\usepackage{float}
\usepackage{array}
\usepackage{tikz}
\usepackage[export]{adjustbox}
\usepackage{subcaption}
\usepackage{enumerate}
\usepackage{pgf}
\usepackage{placeins}

\usepackage{tabularx}
\usepackage{tabulary}
\usepackage{makecell}
\usepackage{todonotes}
\usepackage[mathscr]{euscript}

\usetikzlibrary{arrows,automata}

\newcommand{\kalesha}[1]{{\textcolor[rgb]{0.0,0.75,0.40}{KB: #1}}}

\newcommand{\revision}[1]{{\textcolor[rgb]{0.75,0.0,0.0}{#1}}}  % for general comments/notes/revisions

\DeclareMathOperator*{\argmax}{arg\,max}
\long\def\|*#1*/{}

\title{Active Learning within Constrained Environments through \\Imitation of an Expert Questioner}

\|*
\author{
	Kalesha Bullard\footnote{Contact Author}\And
	Sonia Chernova\\
	\affiliations
	Georgia Institute of Technology\\
	\emails
	ksbullard@gatech.edu,
	chernova@cc.gatech.edu
}
*/

\author{
	Kalesha Bullard\footnote{Contact Author}\and
	Yannick Schroecker\And
	Sonia Chernova\\
	\affiliations
	Georgia Institute of Technology\\
	\emails
	\{ksbullard, yannickschroecker\}@gatech.edu,
	chernova@cc.gatech.edu
}

\|*
\author{
	\affiliations
	\emails
}
*/

\begin{document}
	
\maketitle

\begin{abstract}
	
	Active learning agents typically employ a query selection algorithm which solely considers the agent's learning objectives.  However, this may be insufficient in more realistic human domains.  %it is important to consider external constraints imposed on the learner.  %it is important to additionally consider how time and resource constraints imposed on the agent by its environment should impact its questioning strategy.  
	%This work builds upon prior work in Active Learning for dynamic environments and explores how to enable a learning agent in a constrained environment to adapt its strategy for query selection, by concurrently reasoning about \textit{both} its learning objectives \textit{and} environmental constraints, within its objective function.  Our approach uses Inverse Reinforcement Learning to infer weights for the agent's desiderata by mimicking the strategy of an expert questioner.  Experiments are conducted on a concept learning task to understand the impact of time and resource constraints on the efficacy of solving the learning problem.  Our findings show that the environmentally-aware learning agent always performs \textit{at least} as well as the agents only optimizing for learning objectives, but is able to \textit{statistically outperform} all other active learners explored in a subset of constrained scenarios.  The key implication is autonomous adaptation for active learning agents to more realistic environments, where time and resources are often constrained.
	This work uses imitation learning to enable an agent in a constrained environment to concurrently reason about \textit{both} its internal learning goals \textit{and} environmental constraints externally imposed, all within its objective function.  %Our approach uses Inverse Reinforcement Learning to infer weights for the agent's desiderata by mimicking the strategy of an expert questioner.  
	Experiments are conducted on a concept learning task to test generalization of the proposed algorithm to different environmental conditions and analyze how time and resource constraints impact efficacy of solving the learning problem.  Our findings show the environmentally-aware learning agent is able to statistically outperform all other active learners explored under most of the constrained conditions.  A key implication is adaptation for active learning agents to more realistic human environments, where constraints are often externally imposed on the learner. 
	
	%The approach taken is to infer how an expert questioner trades off the individual learning and environmental decision features, through Inverse Reinforcement Learning.  
\end{abstract}

\section{Introduction}
Active learning (AL) agents are intended to learn from an oracle, often assumed to be human, but typically not \textit{designed} for more realistic human environments. Understanding environmental context however is especially important for robotic agents, generally assumed to be colocated in the environment with the oracle or teacher.  Within the robotics community, there has been AL work aimed at understanding \cite{cakmak2010designing,knox2013training,gonzalez2018analyzing,bullard2018human}, modeling \cite{rosenthal2011modeling,racca2018active}, and improving \cite{chao2010transparent} interaction with a human partner.  An important aspect of the interactive learning problem, this body of work focuses on \textit{interaction} with the teacher, but there still remains the open question of how the learner should integrate reasoning about the \textit{environment} in which it is situated.  

Specifically, external constraints imposed on the learner %may only have an indirect link to the teacher (\textit{e.g.} teacher has limited time or cognitive resources that can be devoted to answering the learner's questions), yet still has 
may have \textit{direct} implications for solving the learning problem. For example, a teacher has only a limited time frame of availability or limited cognitive resources that can be devoted to answering the learner's questions.  This information may need to influence the learner's questioning policy.  However, the problem of trading off learning goals with environmental constraints is relatively unexplored within AL literature, particularly when considering dynamic environments.  Yet this problem is important for learning in realistic human settings.  

In this work, we investigate the question of how to enable an active learner to reason about its learning objectives within a dynamically changing environment while concurrently considering time and resource constraints provided for solving the learning problem.  We use a decision-theoretic approach to active learning, whereby the individual decision criteria (or decision features) within the objective function are hand designed and include both task-centric and environment-centric features. Nonetheless, since the learning agent must consider multiple and diverse decision criteria, it becomes difficult to manually tune the individual objectives.  Thus we propose imitation of an expert questioner for learning to weight the decision features.  Our approach employs Inverse Reinforcement Learning (IRL) for inferring weights of the objective function from demonstrations of an expert policy.  

\begin{figure*}[t]
	\centering
	\includegraphics[width=0.6\textwidth]{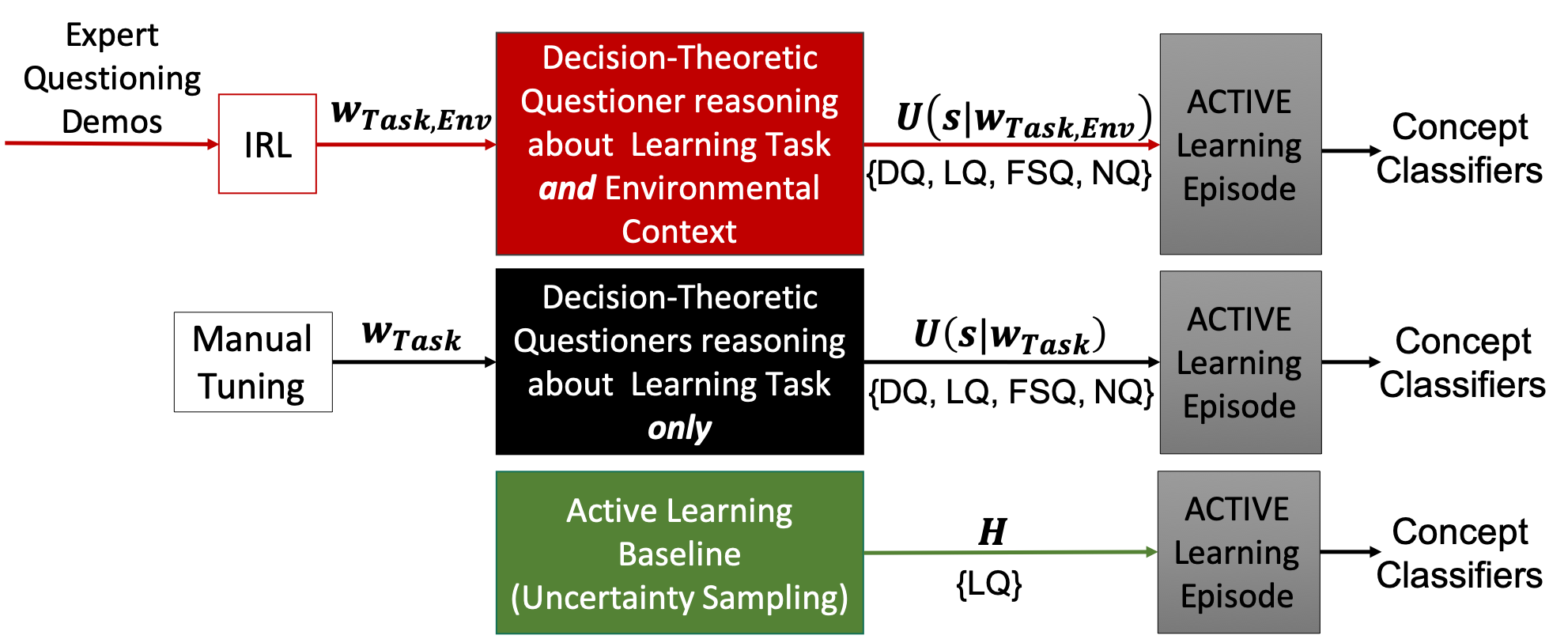}  %, height=5cm
	\caption{Learning system diagram, illustrating how each active learning strategy performs query selection.}
	\label{fig:sys_diagram}
	\vspace{-2mm}
\end{figure*}

In the experiments conducted, the agent is given a concept learning problem that it must use active learning to solve, under different environmentally constrained conditions.  %It is also situated within an environment that dynamically changes in a deterministic yet unknown way.    
%To the best of our knowledge, this work makes the following contributions:  
This work makes the following contributions:
\begin{itemize}
	\item first AL work to reason about environmental constraints within the objective function of the learner
	\item first AL work to use imitation learning for mimicking the policy of an expert questioner
\end{itemize}

%We evaluate the efficacy of our experimental environmentally-aware questioning strategy, as compared to baselines of uncertainty sampling and two task-centric decision-theoretic questioning strategies.  Our findings show the environmentally-aware learning agent always performs \textit{at least} as well as the agents only optimizing for learning objectives, but is able to \textit{statistically outperform} all other active learners explored in most of the constrained scenarios.

We evaluate efficacy using two separate task datasets and show that environmentally-aware reasoning allows our algorithm to significantly outperform an established AL baseline of uncertainty sampling and task-centric questioning strategies examined.

\|*
\revision{
Outline of Thoughts/Points
\begin{itemize}
	\item \textbf{motivation.} motivated by scenario of a robot that exists in an environment with humans, e.g. a personal robot, which aims to be useful by assisting its human partners with tasks. in this context, the robot may be given several tasks, specified by abstract task plans or recipes, for which it must first ground the necessary task-relevant symbols in environment, before it can reason about executing the task.  for example, the robot may be asked to serve breakfast, a task which requires it to first perceptually ground breakfast items in its environment, \textit{e.g.} bread, eggs, and juice.  define the concept/symbol grounding problem \cite{harnad1990symbol} and challenges associated with it for an agent existing in a dynamic environment, who desires to quickly acquire knowledge associated with performing new tasks in its environment.
	\begin{itemize}
		\item environment is changing over time and consequently, concept groundings should be continually adapted/refined in order to sufficiently cover the diversity of samples/instances the agent is expected to later encounter.
		\item as get closer to more realistic settings, teacher will have constraints with respect to the amount of time and cognitive resources he/she can devote to the agent for...well, teaching. so agent needs ability to not only identify its gaps in knowledge towards learning the given concepts (\textit{i.e.} assess how to achieve its learning goals), but also the capability to reason about how to adapts its learning strategy, based upon constraints imposed on it by the teacher or its learning environment (based upon the context of the learning environment in which it is situated).   
	\end{itemize}
	\item \textbf{useful family of approaches.} introduce paradigm of learning from demonstration as a promising approach, given expectation of access to a human partner in the agent's environment. but also primary limitations associated with \emph{passive learning} with naive end users, in dynamic environments: 
	\begin{itemize}
		\item user is required to have proficiency in teaching concepts, if process is to be sample efficient and 
		\item user is required to track state of robot's knowledge in order to aid it in timely and appropriate refinement of its concept models (task knowledge).
	\end{itemize}
	\item \textbf{proposed approach.} propose Active Learning (AL) as the most fitting subclass within larger family of Interactive Machine Learning approaches. define problem of active learning, w/n context of machine learning: form of semi-supervised learning whereby learner queries oracle for labels of unlabeled instances based upon predefined selection criteria (\textit{e.g.} uncertainty). %discuss limitations with current literature: namely, it typically \textit{only} reasons about the learning problem agent must solve \cite{chernova2009interactive,lopes2009active,chao2010transparent,gribovskaya2010active,kroemer2010combining,cakmak2012designing,kulick2013active,daniel2014active,hayes2014discovering,thomason2017opportunistic,basu2018learning,bullard2018towards,racca2018active}, but not time and resource constraints imposed on it by the environment in which the agent is situated.  in other words, in prior literature, the learner typically uses an objective function to reason about its learning goals but does not \textit{additionally} consider environmental constraints. 
	\item \textbf{novel contributions.} concisely state contributions of this work. to the best of our knowledge, this work makes the following contributions:  \\\textbf{\textit{first} AL work to...} 
	\begin{itemize}
		\item (a) reason about environmental context/constraints within the objective function of the learner
		\item (b) use imitation learning for mimicking the policy of an expert questioner
	\end{itemize}
\end{itemize}
}
*/

\|*
\nocite{bullard2018towards}
\nocite{cakmak2012designing}
\nocite{chao2010transparent}
\nocite{chernova2009interactive}
\nocite{daniel2014active}
\nocite{gribovskaya2010active}
\nocite{hayes2014discovering}
\nocite{kroemer2010combining}
\nocite{kulick2013active}
\nocite{lopes2009active}
\nocite{racca2018active}
\nocite{racca2019teacher}
\nocite{thomason2017opportunistic}
*/

\section{Related Work}
%\revision{should probably have subsections on both Active Learning and Inverse Reinforcement Learning, given the contributions claimed.  will also need to look up most recent references for AL. maybe also need to extend to AL for AI more broadly, not just robotics.}

Active Learning encompasses an extensive body of literature, spanning across several problem domains.  We focus here on the most relevant work within the broader space, active learning for robots and embodied artificial agents.  Most literature in AL for robots solves learning problems directly relevant to robotics domains: learning an expert policy to derive desirable robot behavior \cite{chernova2009interactive,lopes2009active,kroemer2010combining,cakmak2012designing,daniel2014active,basu2018learning}, inferring sequencing constraints on actions in a task \cite{hayes2014discovering}, and grounding task-relevant symbols or descriptions \cite{chao2010transparent,kulick2013active,thomason2017opportunistic,bullard2018towards}.  A key limitation however is current approaches \textit{only} reason about the learning problem the agent must solve, but not time and resource constraints imposed by the environment in which the agent is situated.  In other words, in prior literature, the learner typically uses an objective function to reason about its learning goals but does not \textit{additionally} consider environmental constraints. 
%\revision{Primary limitation with current literature: namely, it typically \textit{only} reasons about the learning problem agent must solve, but not time and resource constraints imposed on it by the environment in which the agent is situated.  In other words, in prior literature, the learner typically uses an objective function to reason about its learning goals but does not \textit{additionally} consider environmental constraints.}  
%As noted earlier, there is also a small community of AL literature that focuses on the interaction aspect, but this is not directly relevant, as we make no assumptions about the teacher beyond the given constraints possibly deriving from teacher preferences.  

The most relevant prior work explored autonomous arbitration between multiple types of active learning queries, acquiring both feature and instance input from the teacher, and situated in dynamic environments \cite{bullard2018towards}.  A primary contribution of this work was an algorithm for arbitration between diverse query types that could also adapt the frequency of its questioning to the rate at which objects changed in its environment.  Nonetheless, a key limitation is the inability to reason about constraints imposed by the teacher or learning environment.  Thus, while able to adapt its questioning strategy to the rate of environmental change, the learner has no mechanism for adapting its strategy to the query budget given or amount of learning time allocated.

\section{Problem Formulation and Approach}

Symbol (or concept) grounding is the problem of mapping symbolic representations to constructs in the physical world \cite{harnad1990symbol}.  Specifically, the agent must solve a \textit{task-situated} concept grounding problem,  %\cite{bullard2016grounding}.
whereby it is given abstract task-relevant concepts to be perceptually grounded in its environment, in a way appropriate for the task.  For example, when learning concepts for the \textit{serve breakfast} task, \textit{eggs} scrambled or sunny-side up would be a more appropriate grounding than a dozen cartoned eggs. 

We formalize the problem as follows: Given a set of objects $\textbf{X}$ from a scene in the agent's purview taken at time $t$, each object instance $\textbf{x} \in \textbf{X}$ is represented by a feature vector $\boldsymbol{x^{t}} = <f_{1}^t . . . f_{m}^t>$.  We assume the agent has both exteroceptive and proprioceptive sensors for perceiving its external environment and internal state.  Each object instance then is modeled by the superset of features $F$ extracted from the agent's sensors at $t$ (\textit{e.g.} object height or color, position of robot base or end effector).  %and derived by computation, {\textit{e.g.} position of object with respect to an interest point}.  
A set of binary classifiers, one for each symbol $y \in Y$, the set of object symbols, each take as input an instance $\textbf{x}$ and produce a degree of confidence $p(y|\textbf{x}) = [0,1]$ that $\textbf{x}$ has label $y$. For each symbol, a Gaussian Process Classifier was trained.  % using a standard radial basis function kernel; 
This representation was selected because it both probabilistically models agent uncertainty and learns well from sparse data.  %as it both provides probabilistic output and learns well from sparse data.

In dynamic environments, groundings also change over time and concept models must be refined accordingly.  This may include change in the physical object state (\textit{e.g.} eggs going from being in a shell to scrambled) or objects being replaced within the same category (\textit{e.g.} breakfast beverage being served one day as coffee in a mug and another as orange juice in a glass).  %In order for a human partner to help the agent appropriately refine its grounded concept models, the person must both be proficient at teaching and track the agent's knowledge over time.  This is not a reasonable expectation for naive users and motivates the use of active learning for the agent to direct its own learning process. 
Since it is unreasonable to expect a human partner to track the agent's knowledge over time, in a changing environment, we take an \textit{active} concept grounding approach.  
%Additionally, the teacher is able to provide different types of input towards enabling learning, namely informative features for representing concepts and training instances.  

\subsection{Active Learning for Concept Grounding}

%Prior work has explored autonomous arbitration between multiple types of active learning queries, acquiring both informative features and representative training instances from a teacher, in dynamic environments \cite{bullard2018towards}.  
%Our approach builds upon this prior work and contributes the \textit{addition} of reasoning about environmental constraints.  Specifically, we examine the imposition of external time and resource constraints.

Active Learning enables a learner to query an oracle or teacher for information about which is has uncertainty.  It typically assumes a query will be made at every turn and seeks to equip the learner with a utility function for selecting an \textit{optimal} query \cite{settles2012active}.  However, real world environments often do not allow the learner unlimited queries or time for querying, and simultaneously change over time.  This means it is not always the best use of time and resources to make a query at every time step, until the query budget is depleted.  Thus, employing a traditional AL strategy may not maximize learning in dynamically changing, constrained environments, as shown by \cite{bullard2018towards}.  

We present a decision-theoretic AL approach which extends prior work intended for dynamic environments \cite{bullard2018towards}.  
In that work, the authors contributed a decision-theoretic framework for arbitrating between multiple types of AL queries, acquiring both informative features and representative training instances from a teacher.  Building upon that framework, our approach contributes a model that is able to reason about \textit{both} the agent's concept learning goals \textit{and} external time and resource constraints imposed on the agent.  
Specifically, the objective function of the learner is \textit{expanded} to include decision criteria which reason about environmental context.  Equation \ref{eq:expected_utility_action} shows the learner's objective function used at each turn $t$ to assess the expected utility (EU) of an action $a$, given the current learning state $s_t$.

The learning state at $t$ includes: \{estimate of posterior probability distributions of $y \in Y$ for all $\textbf{x} \in \textbf{X}$, interaction history, query budget, and teaching time allocation\}.  
The set of candidate actions $A_t$ consist of demonstration queries for each of the task-relevant concepts $\left[DQ(y) \; \forall y \in Y\right]$, label queries for each object in the current scene $ \left[LQ(\textbf{x}) \; \forall \textbf{x} \in \textbf{X}\right] $, a feature subset query $\left[FSQ\right]$ to identify relevant features for discriminating between task concepts, and a no query action $\left[NQ\right]$.  Thus, there are $|A_t| = |Y| + |\textbf{X}| + 2$ candidate actions from which the agent can choose at each turn $t$.  Additionally, each of the query types is associated with a cost, given a priori.  The learner selects an optimal action $a^*$ as 
\begin{equation} \label{eq:expected_utility_action}
	\begin{split}
		a^* &= \argmax_{a} \; EU(a | s_t) \\
		&= \argmax_{a} \; \sum_{s_{t+1}} \; P(s_{t+1} | a, s_t) \enspace U(s_{t+1}) 
	\end{split}
\end{equation}   
where 
\begin{equation} \label{eq:utility_state}
U(s) = w_{1} \phi_{1}(s) + w_{2} \phi_{2}(s) + ... + w_{n} \phi_{n}(s)
\end{equation}

The set of decision features $\phi \in \Phi$ used in computing $U(s)$ comprise the representation for the agent's objective (decision) function and is primarily what distinguishes prior work from the approaches introduced in this work.  $U(s)$ is represented as a function of decision features $\phi: S \rightarrow [0,1]^k$, where $k$ is number of decision criteria or individual objectives for which the agent is optimizing.   
%In addition to the uncertainty sampling baseline, we employ a decision-theoretic approach from prior literature, as a baseline for comparison.  

\subsubsection{Baseline Approaches}
%Because of this, we employ one traditional AL approach as a baseline for comparison.  
We employ two AL models from prior literature, as baselines for comparison: a standard uncertainty sampling approach (\textbf{U-sampling}) and a state-of-the-art decision-theoretic approach for arbitrating between diverse query actions (\textbf{DT-iros}).   
Uncertainty sampling algorithms are possibly the most commonly employed class of AL strategies in the literature \cite{settles2012active,fu2013survey}.  They assume a single hypothesis $\theta$ and utilize the posterior probability distribution over labels $y \in Y$ given unlabeled instance $x$, $p_{\theta}(Y|\textbf{x})$, in order to detect outliers or instances closest to a decision boundary.   %Uncertainty sampling algorithms optimize for uncertainty, with respect to the predicted class label.  
Like other standard AL approaches, they query at every turn, each time requesting a label for a maximally informative instance, based upon predetermined selection criteria. 
A commonly used metric for uncertainty sampling is \textit{prediction entropy}: $ - \sum_{y \in Y} p_{\theta}(y|\textbf{x}) \log  p_{\theta}(y|\textbf{x}) $.  We employ this as our standard AL baseline (\textbf{U-sampling}).  %Of the action types introduced earlier (demo queries, label queries, feature subset query, no query), this algorithm only reasons about label queries, as this is consistent with standard AL algorithms.  

%\kalesha{Should continue about baseline DT approach (iros) here....  }
Decision-theoretic approaches to active learning simulate all possible outcomes of each candidate query action and optimize with respect to future \textit{expected} utility.  This work builds from prior work employing decision theory to arbitrate between diverse types of learning queries, including a supplemental no-query action \cite{bullard2018towards}.  
The set of decision features investigated were \textit{average classifier discriminability} and \textit{class distribution uniformity}.  

Given a set of instances in the agent's purview $\textbf{X}$ and a task-relevant concept $y$, the \textit{classifier discriminability} metric assesses the \textit{range} of probabilities over the set of instances: $p_\theta(y | x_{max}) - p_\theta(y | x_{min})$, where $x_{max}$ and $x_{min}$ are the model's prediction of the most and least probable examples of class $y$, respectively.  Range is a standardized metric of statistical dispersion; an average is taken over all $y \in Y$.  %Intuitively, this metric incentivizes the learner for differentiating between the most prototypical and improbable examples of each class.  
Class distribution uniformity assesses selection bias in the training sample, due to an unrepresentative class distribution.  It is a useful decision feature in sparse data environments, as has traditionally been the assumption in Learning from Demonstration settings, where the learner does not have sufficient evidence to confidently infer the underlying distribution of classes.  This metric incentivizes the learner to minimize sample selection bias.  
%It is a useful decision feature in sparse data environments, since the learner does not have sufficient data to automatically infer the underlying distribution of classes.  It incentivizes the learner to collect a training  data sample with a uniform distribution of classes, given that there is no reason to assume bias towards any task-relevant class %unrepresentative of the underlying distribuiton of classes., 
%as is typically assumed in LfD environments since the human teacher must provide each training example.  
Given this work also seeks to arbitrate between \textit{all} action types, we employ this previously published decision-theoretic objective function, where the number of decision features $k=2$, as our state-of-the-art baseline (\textbf{DT-iros}).

\|*
, as defined by the following equations: $$ACD(s) = \dfrac{1}{|Y|} \sum_{y \in Y} \left[ p_s(y | o_{max}) - p_s(y | o_{min}) \right] $$  
$$CDU(s) = \dfrac{|D_{y_{min},s}|}{|D_{y_{max},s}|} $$ 

where $p_s$ represents the probabilistic prediction for $y$ in the current state $s$, and $o_{max}$ and $o_{min}$ represent the objects in the scene predicted to be the \textit{most} and \textit{least} probable examples of symbol $y$, respectively, and $D_{y_{max},s}$ and $D_{y_{min},s}$ are each subsets of the training sample.  They represent the sets of positive examples for $y_{max}$ and $y_{min}$, the symbols most and least represented in the training sample at state $s$.  
*/
\|*
Uncertainty sampling was selected as the AL baseline, as it is commonly employed in AL literature \cite{settles2012active,fu2013survey}.  Of the action types introduced earlier (demo queries, label queries, feature subset query, no query), this algorithm only reasons about label queries, as this is consistent with most standard AL algorithms.  It uses entropy to estimate uncertainty of each candidate label query.  
\|*, computed as follows: 
\begin{equation*} \label{eq:entropy}
x^* = \argmax_{x} \;  H(y|x) =  \argmax_{x} \; \left\lbrace - \sum_{y \in Y} P_\theta (y | \textbf{x}) log P_\theta (y | \textbf{x}) \right\rbrace
\end{equation*}
where $H(y|x)$ is the entropy associated with class label $y \in Y$, given instance $x$.  
The decision-theoretic baseline previously published (DT iros) in \cite{bullard2018towards} was designed to arbitrate between diverse types of queries and thus can reason about all four action types.  As DT-task and DT-task-env, both introduced in this work, build from this prior work, they too are able to reason about all four query types.  Figure \ref{fig:sys_diagram} visually summarizes the overall learning system.  Here, $w_{Task}$ and $w_{Task,Env}$ represent the decision feature weight vectors for DT-task and DT-task-env respectively.
*/
%The decision-theoretic learners can all arbitrate between all action types.

\subsubsection{Experimental Approaches}
We introduce two experimental questioning policies: a learning-centric model intended to improve the state-of-the-art (\textbf{DT-task}) and an environmentally-aware active learner (\textbf{DT-task-env}).  
For the learning-centric model, we propose two \textit{additional} decision features that we believe improves the performance of the originally published DT-iros algorithm, even before consideration of environmental context: \textit{instance variation} and \textit{label prediction margin}, defined by Equations \ref{eq:instance_variation} and \ref{eq:prediction_margin} respectively.  %\kalesha{need to figure out what notation is best for these equations and possibly modify accordingly?}

\begin{equation} \label{eq:instance_variation}
	IV(s) = \frac{1}{|Y|} \sum_{y \in Y} \; \frac{\sigma(p(\textbf{X}|y))}{\mathbb{E}[p(\textbf{X}|y)]}
\end{equation}

\begin{equation} \label{eq:prediction_margin}
	PM(s) = \frac{1}{|X|} \sum_{x \in X} \; p_{\theta_1} (y_1 | \textbf{x}) - p_{\theta_2} (y_2 | \textbf{x})
\end{equation}

\textit{Instance variation} is a standardized measure of statistical dispersion.  %, along a different dimension.   
Given a class $y$ and a set of scene instances $\textbf{X}$, it is a measure of relative standard deviation of the class conditional distribution $p_\theta(\textbf{X}|y)$.  Intuitively, it attempts to assess each classifier's ability to recognize variation \textit{amongst} the set of unlabelled instances.  %This metric incentives the selection of queries which increase the learner's \textit{recognition} of the underlying diversity that exists within the unlabelled pool of instances.  

In the context of concept learning, the class-conditional distribution $p(\textbf{X}|y)$ can be thought of as the likelihood of each unlabelled instance $\textbf{x} \in \textbf{X}$ being selected as an example of class $y$.  %The underlying distribution is unknown, since it can depend on a number of latent factors (\textit{e.g.} teacher preferences, how instances are generated or populated in the environment), but also assumed to be multi-modal, 
Given that multiple, diverse instances within a scene may serve as positive examples of a given class, it seems useful to employ decision features which approximate the learner's ability to recognize diversity amongst the set of unlabelled instances in its purview.  
Because of this, both \textit{classifier discriminability} and \textit{instance variation} are measures of statistical dispersion, but along different dimensions.  Whereas, classifier discriminability is a measure of statistical dispersion over the likelihood of instances belonging to a class, instance variation quantifies the statistical dispersion over the features values of instances.  The former rewards the learner for differentiating between the most prototypical and improbable examples of each class; the latter rewards the learner for recognizing greater variation between instances.  
Both decision features incentivize the selection of queries which increase the learner's \textit{recognition} of the underlying \textit{diversity} that exists within the pool of unlabelled instances.  %Both decision features incentivize \textit{diversity} in the training samples being acquired. 

Given an unlabelled instance $x$ and a distribution over class labels $p(\textbf{Y}|x)$, \textit{label prediction margin} measures the difference between what the learner predicts to be the most probable label $y_1$ and second most probable label $y_2$.  Previously employed in AL literature \cite{settles2012active}, it is a measure of uncertainty; as the margin increases, the learner is more confident about its prediction.  It is computed for all scene instances, then averaged.  This decision feature incentivizes \textit{accuracy} in the class prediction for each unlabelled instance.

Thus, the first decision function proposed in this work (\textbf{DT-task}) subsumes the set of decision features considered by DT-iros, considering \textit{four} learning-centric criteria that each optimize for different aspects of the concept learning problem.  

The primary contribution of this work however is in the addition of \textit{environmental context} into the AL agent's objective function.  We \textit{introduce} the following environmental features:  %\kalesha{do i need to write an equation for the first two as well?}
\begin{itemize}
	\item \textit{query budget consumption} -- measures the proportion of query budget consumed at turn $t$, given the query history
	\item \textit{remaining time usage} -- measures the proportion of allocated time remaining after turn $t$
	\item \textit{non-query time passed} -- measures the proportion of consecutive turns no query was made within a sliding time window $T_w$; here the size of the time window is proportional to the rate of environmental change; it is computed as $ t_{NQ}=\frac{n_{NQ}}{|T_w|}$; $t_{NQ}\rightarrow 0$ when the learner has just queried and $t_{NQ} \rightarrow 1$ when the learner has \textit{not} queried throughout the entire duration of the time window; intuitively this metric is intended to penalize the agent for being \textit{too} passive, in a dynamically changing environment
\end{itemize}

The environmentally-aware agent's objective function (\textbf{DT-task-env}) is then composed of a linear combination of \textit{seven} decision features, a subset of which roughly attempt to estimate progress towards learning goals (\textit{i.e.} learning \textit{task} centric) and the remaining features intended to incentivize wise time and resource management (\textit{i.e.} \textit{environment} centric).  All decision-theoretic learners described can arbitrate between \textit{all} communicative action types.  

Given the different types of decision features being considered however, it is challenging to decipher how to trade them off (\textit{e.g.} budget consumption versus prediction margin).  One key observation is humans can often intuitively reason about decision criteria that are difficult to compare quantitatively; thus, we propose to observe the strategy of a human expert questioner, given the same learning problem, and infer how the expert trades off the given decision criteria.

\|*
\revision{Write here about}...active learning formulation for solving concept grounding problem, built from prior work on using decision-theoretic AL in dynamic environment \cite{bullard2018towards}. should discuss state space, candidate actions, and decision criteria to be considered by agent.
*/

\subsection{Imitating an Expert Questioning Strategy}

Imitation learning seeks to efficiently learn desired behavior by mimicking a domain expert \cite{osa2018algorithmic}.  Within imitation learning literature, Inverse Reinforcement Learning (IRL) aims to recover the expert's reward (objective) function from demonstrations of a policy \cite{ng2000algorithms,abbeel2004apprenticeship}.  We employ a state of the art IRL algorithm, maximum entropy IRL \cite{ziebart2008maximum}, to infer the weights $w \in W$ of Equation \ref{eq:utility_state}, for the active learner's decision features, $\forall \phi \in \Phi$, as wielded by an expert.  

The maximum entropy loss function $L_{ME}$ maximizes entropy of distributions over paths followed by the expert, while satisfying the constraint that the learner's decision feature counts should ideally match those of the expert.  %\revision{In this context, feature counts are ...}
The problem is formulated as follows: Given  $\phi_E(\tau) \;\; \forall \tau \in T^{demo}$, find an optimal weight vector $\textbf{w}$ such that  
\begin{equation} \label{eq:max_ent_irl_formulation}
	\textbf{w}^{*} = \argmax_{\textbf{w}} \;\; - \sum_{\tau} \; p(\tau | \textbf{w}) \; \ln p(\tau | \textbf{w})
\end{equation}

subject to the constraint  
\begin{equation} \label{eq:feature_matching_constraint}
	\mathbb{E}\left[\phi_E(\tau)\right] = \mathbb{E}\left[\phi_L(\tau)\right]
\end{equation}

where $\phi_E(\tau)$ and $\phi_L(\tau)$ represent the feature counts of the expert and learner respectively, for a trajectory $\tau$.  % ... feature expectations for each expert trajectory $\tau \in T^{demo}$, the set of expert demonstrations.  
In our problem domain, a trajectory is a sequence of learning states visited and communicative actions taken $\{s_1,a_1,s_2,a_2, ... s_T,a_T\}$ at each time step $t <= T$, the maximum number of iterations allowed in a learning episode.

Optimization using the maximum entropy loss $L_{ME}(\textbf{w})$ is equivalent to maximizing the log likelihood of the expert demonstrations \cite{ziebart2008maximum,osa2018algorithmic}: %, as shown in the derivation below.
\begin{equation*} \label{eq:max_ent_irl_loss}
	\begin{split}
		\textbf{w}^{*} & = \argmax_{\textbf{w}} \;\; L_{ME}(\textbf{w}) \\
		& = \argmax_{\textbf{w}} \;\; \sum_{\tau} \; p(\tau | \textbf{w}) \; \ln \frac{1}{p(\tau | \textbf{w})} \\
		& \propto  \argmax_{\textbf{w}} \;\; \sum_{\tau} \; p(\tau | \textbf{w}) \\
		\textbf{w}^{*} & \propto  \argmax_{\textbf{w}} \;\; \sum_{\tau} \; \ln p(\tau | \textbf{w})
	\end{split}
\end{equation*}

%where $L_{ME}(\textbf{w})$ represents the maximum entropy loss given a set of objective function weights $\textbf{w}$.  

Using this formulation, the gradient of the IRL loss, shown in Equation \ref{eq:max_ent_irl_gradient},  %with respect to the decision feature weights 
is the difference between the empirical feature counts (demonstrated by the expert) and the expected feature counts, computed from sample trajectories generated with \textbf{w}.

\begin{equation} \label{eq:max_ent_irl_gradient}
	\nabla_{\textbf{w}} L_{ME} = \mathbb{E_{\pi^E}}\left[\phi_E(\tau)\right] - \sum_{\tau} p(\tau | \textbf{w}) \phi_L(\tau)
\end{equation}

We used an empirically determined maximum number of iterations as stopping criteria for the IRL algorithm. Weights for the environmentally-aware active learner's objective function were learned offline and tested for generalization in AL episodes under different environmental conditions.  %, then tested for efficacy in solving a given learning problem under different environmentally constrained conditions.  

\|*
Write here about...learning the agent's objective function for trading off learning goals and environmental constraints, using Inverse RL \cite{ziebart2008maximum}. \revision{cite other IRL papers for this, particularly the seminal papers in the field}. motivate why this is necessary.
*/

\subsection{Learning Episode}
%\revision{describe traversal of learning episode.} 

Figure \ref{fig:sys_diagram} shows the high-level flow for the learning system.  For each active learning episode conducted, the task-relevant concepts and questioning strategy are given as input.  
Within an episode, at each turn $t$, the agent perceives all objects in its purview, computes its estimate of the posterior probability distributions $p(y|x) \; \forall x,y$ to update learning state $s_t$, determines the set of candidate actions $A_t$, computes $EU(a | s_t) \; \forall a \in A_t$, then takes an optimal action $a^*$. The learning episode concludes once $t = T$.

\|*
\subsection{Not Quite Done...}
\textit{more} major points to discuss:
\begin{itemize}
	%\item introduce concept learning problem to be solved. 
	%\item active learning formulation for solving learning problem, built from prior work on using decision-theoretic AL in dynamic environment. should discuss state space, candidate actions, and decision criteria to be considered by agent.
	%\item learning the agent's objective function for trading off learning goals and environmental constraints, using Inverse RL \cite{ziebart2008maximum}. \revision{cite other IRL papers for this, esp the seminal papers in the field}. motivate why this is necessary.
	\item describe traversal of learning episode. 
\end{itemize}
*/

\section{Evaluation}
This work explores an AL strategy \textit{designed} to optimize for environmental constraints
%The AL algorithm being investigated in this work is the first to explore an objective function designed to optimize the learner's query selection strategy for constraints imposed by its learning environment 
and proposes an imitation learning approach for accomplishing this.  Toward this end, we test two hypotheses: (1) Reasoning \textit{additionally} about environmental context can enable an AL agent to adapt its questioning strategy and improve its learning performance under constrained conditions, and (2) Imitation Learning can be used to infer an expert's strategy for managing time and resources allocated to solve a given learning problem, then generalized to other constrained environments.  
As illustrated in Figure \ref{fig:sys_diagram}, each of the questioning strategies conduct their own learning episodes, during which binary classifiers are trained for all task-relevant classes, based upon information gathered.  
We use recognition accuracy on hold-out test sets for assessing the concept models learned, given each questioning strategy.

\subsection{Experimental Design}
In evaluating the AL approaches, we focus experiments on a concept grounding task in a dynamic environment, under \textit{different} environmentally constrained conditions.  We also examine performance on another task for generalization of the learned decision feature weights across tasks.  Both concept grounding tasks are given the same four abstract concepts to ground (main dish, snack, fruit, and beverage), but are generated from different object RGB-D datasets and represent different properties of dynamic change.  %For each task, learning curves are averaged over 5 runs, each run sampling from a different training dataset. The training datasets are all generated from larger RGB-D datasets, described in detail in

\begin{figure}[tb]
	\centering
	\begin{adjustbox}{minipage=\linewidth,scale=0.85}
	\centering
	%\vspace{3mm}
	\begin{subfigure}[b]{0.20\columnwidth}  %\columnwidth
		\centering
		\includegraphics[width=\textwidth]{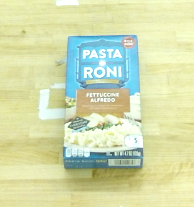}
		\caption{Pasta \\(\textit{state}:in box)}
		\label{fig:pasta_state1}
	\end{subfigure}	
	\begin{subfigure}[b]{0.22\columnwidth}
		\centering
		\includegraphics[width=\textwidth]{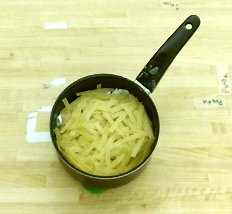}
		\caption{Pasta \\(\textit{state}:in pot)}
		\label{fig:pasta_state2}
	\end{subfigure}	
	\begin{subfigure}[b]{0.22\columnwidth}
		\centering
		\includegraphics[width=\textwidth]{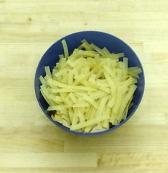}
		\caption{Pasta \\(\textit{state}:in bowl)}
		\label{fig:pasta_state3}
	\end{subfigure}
	\\
	\|*
	\begin{subfigure}[b]{0.24\textwidth}
		\centering
		\includegraphics[width=\textwidth]{figures/snack_2_2_2.png}
		\caption{Snack}
		\label{fig:pasta_state4}
	\end{subfigure}
	*/
	\begin{subfigure}[b]{0.20\columnwidth}
		\centering
		\includegraphics[width=\textwidth]{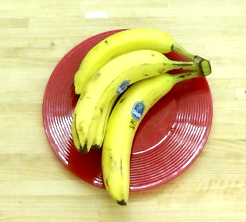}
		\caption{Banana \\(\textit{state}:bunch)}
		\label{fig:pasta_state5}
	\end{subfigure}
	\begin{subfigure}[b]{0.20\columnwidth}
		\centering
		\includegraphics[width=\textwidth]{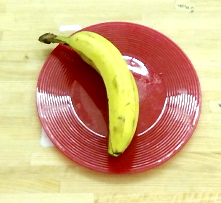}
		\caption{Banana \\(\textit{state}:single)}
		\label{fig:pasta_state6}
	\end{subfigure}
	\begin{subfigure}[b]{0.20\columnwidth}
		\centering
		\includegraphics[width=\textwidth]{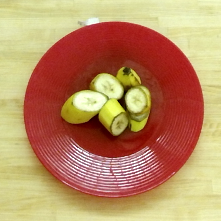}
		\caption{Banana \\(\textit{state}:sliced)}
		\label{fig:pasta_state7}
	\end{subfigure}
	\|*
	\begin{subfigure}[b]{0.24\textwidth}
		\centering
		\includegraphics[width=\textwidth]{figures/beverage_1_2_1.png}
		\caption{Beverage}
		\label{fig:pasta_state8}
	\end{subfigure}
	*/
	
	\end{adjustbox} 
	\caption{ Illustration of object state changes for \textit{main dish} and \textit{fruit} objects classes in \emph{prepare-lunch} task.}\label{fig:prepare_lunch_objs} %Juxtaposed against \textit{snack} and \textit{beverage} object classes whereby the states of the objects remain static.  \revision{\\Crop images so that objets are easier to see.}
	\vspace{-3mm}
\end{figure}

%\kalesha{In this paragraph, apparently should clarify that the training datasets are all generated from larger RGB-D datasets, described in detail in [Bullard et al, 2018b]? (see rebuttal to R3).}  
The \textit{prepare-lunch} task, the most difficult of the two learning problems, is our focus; it places emphasis on the same objects changing \textit{state}, as one might expect over the course of the task (\textit{e.g.} pasta going from being in a box in the pantry to being cooked in a pot to being served in bowl for lunch.)  It was extracted from a local RGB-D object dataset focused on state-change.  Figure \ref{fig:prepare_lunch_objs} shows an example of the type of dynamic change the learner may expect to see in this task.  In the second \textit{pack lunchbox} task, objects do not change state, but have greater within-category diversity.  For example, the \textit{fruit} class contains apples, oranges, peaches, and pears, and the \textit{beverage} class contains varieties of both soda and water.  This task was extracted from the
University of Washington RGB-D dataset of common household objects \cite{lai2011large}.  As both tasks are from prior literature, details regarding data collection can be found in \cite{bullard2018towards}.

% Image of task demos

%We conducted our evaluation on two different concept grounding tasks, each given the same four abstract concepts to ground (main dish, snack, fruit, and beverage), but generated from two different object RGB-D datasets and representing different properties of dynamic change.  Most of our evaluation was 

% Image of task demos
\begin{figure*}[tb]
	\centering
	%\vspace{3mm}
	%\begin{adjustbox}{minipage=\linewidth,scale=1.0}
	\begin{subfigure}[b]{0.35\textwidth}  %\columnwidth
		\centering
		\includegraphics[width=\textwidth]{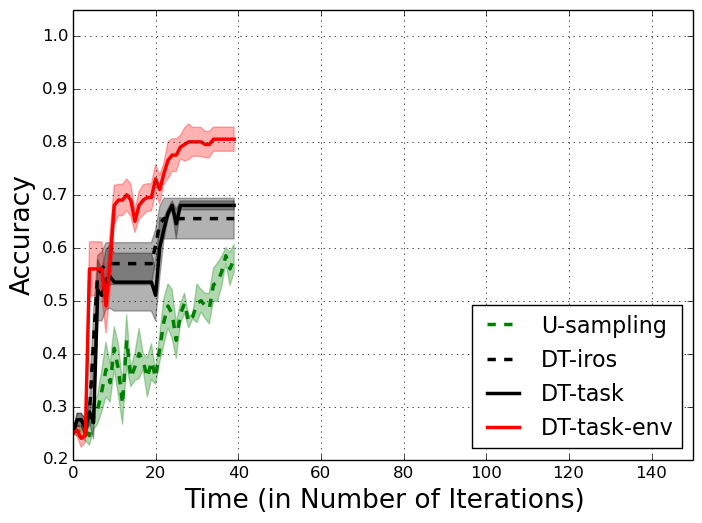}
		\caption{CONDITION 3: Constrained \textit{Time}}
		\label{fig:prepare_lunch_experiment3}
	\end{subfigure}	
	\begin{subfigure}[b]{0.35\textwidth}
		\centering
		\includegraphics[width=\textwidth]{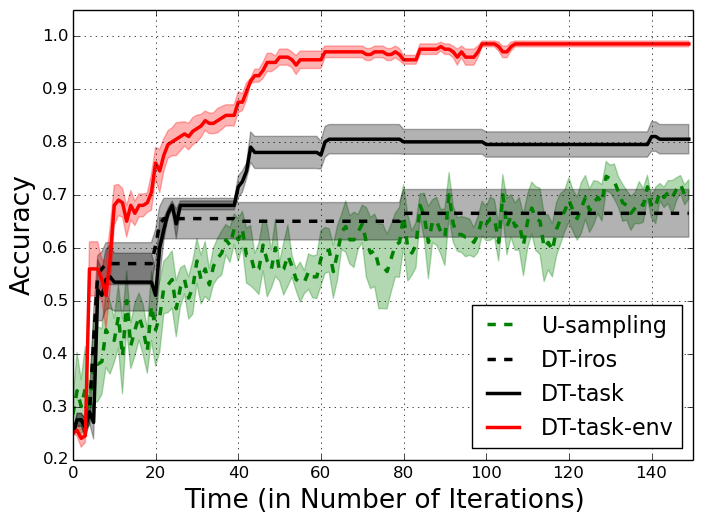}
		\caption{CONDITION 4: Unconstrained}
		\label{fig:prepare_lunch_experiment4}
	\end{subfigure}	
	\begin{subfigure}[b]{0.35\textwidth}
		\centering
		\includegraphics[width=\textwidth]{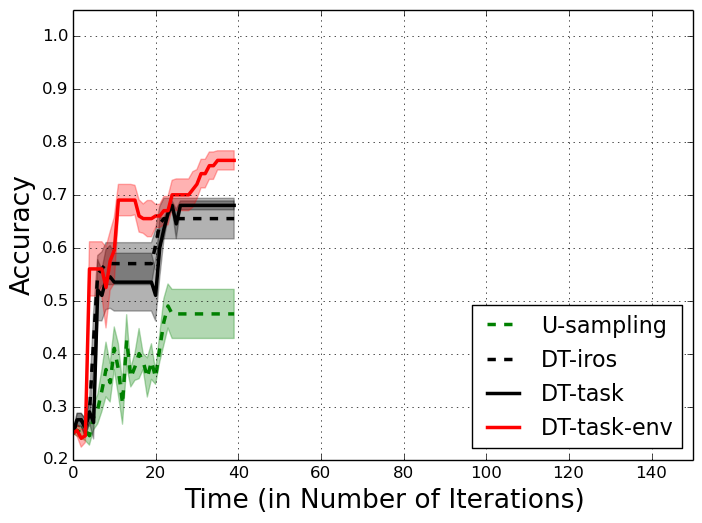}
		\caption{CONDITION 1: Constrained \textit{Time} \& \textit{Budget}}
		\label{fig:prepare_lunch_experiment1}
	\end{subfigure}
	\begin{subfigure}[b]{0.35\textwidth}
		\centering
		\includegraphics[width=\textwidth]{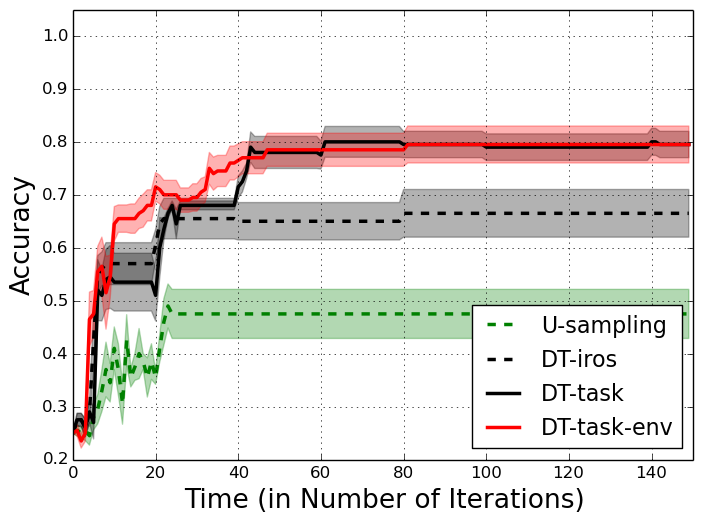}
		\caption{CONDITION 2: Constrained \textit{Budget}}
		\label{fig:prepare_lunch_experiment2}
	\end{subfigure}
	%\end{adjustbox}
	\caption{Prepare Lunch Task. Shows performance (test accuracy with standard error) for each AL strategy under different environmentally constrained conditions. Parameters of allocated time and query budget imposed on the learner vary, with: (a) \textit{only time} constrained [budget: high (500), time: low (40)], (b) \textit{neither} time nor query budget constrained [budget: high (500), time: high (150)], (c) \textit{both} time and query budget constrained [budget: low (25), time: low (40)],  and (d) \textit{only} query \textit{budget} constrained [budget: low (25), time: high (150)]. }\label{fig:prepare_lunch_learning_curves} %\kalesha{Write a caption here, briefly describing the four experiments for \textit{serve pasta task}. And FIX sub-caption formatting.  Explain what is being \underline{defined} as \textbf{\textit{constrained}} versus \textbf{\textit{unconstrained}}, so as to minimize questions.}
	\vspace{-3mm}	 	 
\end{figure*}

Since RGB-D datasets were being used for evaluation, in order to create a learning environment that more closely approximates real-world settings, we simulated multi-modal features, representing features extracted from a robot's other sensors\footnote{\textit{object's location relative to interest points in the environment (\textit{e.g.} counter top, stove, refrigerator, pantry), the object`s location relative to the robot base, absolute location of robot's base in the environment, location of the robot`s base with respect to the counter top, the robot`s joint positions for each arm, pose of the robot's hands, robot`s hand states (open vs closed), weight of the object, and max/min/average volume of noise in the environment over duration of learning episode}}.  Gaussian noise was added to all simulated features since robot sensor data is typically noisy.  We also simulated dynamic change in the environment by sampling a new set of object images at a predetermined rate, to represent the scene changing.  %, where objects may change state.  %, like illustrated in Figure \ref{fig:prepare_lunch_objs}.  
At each turn $t$, $O$ contains only one observation (image) of each object in the scene.  To simulate environmental change, the perceptual system generates a new set of observations.  Else, it outputs the set of observations from $t-1$.  In the prepare-lunch task, objects can change state.  
%With the scene change, objects may change state, like illustrated in Figure \ref{fig:prepare_lunch_objs}.  

Using the complete RGB-D object datasets, we generated five smaller \textit{task} training data samples and one disjoint hold-out test sample for each task.  Since the UW dataset is several orders of magnitude larger than the local dataset created, the data sample sizes vary by task.  Each of the training and test task samples are 80 images and 40 images respectively for the prepare-lunch task and 3200 images and 800 images respectively for the pack-lunchbox task. %The training and test data samples generated are disjoint.

There are four AL algorithms being evaluated on each task: the uncertainty sampling baseline (\textbf{U-sampling}), two task-centric decision-theoretic learners (\textbf{DT-iros} and \textbf{DT-task}), and our experimental environmentally-aware decision-theoretic approach (\textbf{DT-task-env}), which reasons about \textit{both} learning objectives \textit{and} environmental constraints.  
%The task-centric approaches (DT-iros and DT-task) only include decision criteria pertaining to learning objectives.  The environmentally-aware approach (DT-task-env), learned through IRL, reasons about \textit{both} learning objectives \textit{and} environmental constraints.  All of these decision theoretic approaches, primarily intended for learning in dynamically changing environments, are compared against an uncertainty sampling based active learner (U-sampling), as this is a standard approach in AL literature \cite{settles2012active,fu2013survey}.  It uses entropy to estimate uncertainty of each candidate instance query, computed as follows:

For training of DT-task-env decision feature weights, an expert questioner was given a very constrained query budget (15) and time allocation (30 turns) to ground the \textit{prepare-lunch} task concepts.  During training, the environmental scene changed every 10 turns, and major object state changes took place with each scene change; thus spreading the query budget out over the allocated time period affords the opportunity to acquire a more diverse and representative training sample.  This was a key part of the strategy employed by the expert used, which was one of the authors of this paper.  

The expert provided three demonstrations of questioning sessions (learning episodes) in the training scenario.  
As part of the strategy demonstrated, the expert also always requested relevant features for discriminating between concepts (FSQ) early in the learning episode (within the first five turns), focused most queries on the least costly query type, and focused on quickly acquiring representative training examples for each class.  During IRL training, the maximum number of iterations was set to 100, and we selected the set of weights $\textbf{w}^*$ that performed best on the validation set.  Qualitatively examining the rollouts associated with $\textbf{w}^*$ under the training conditions, the behavior of the imitation learner was able to closely match the expert's questioning strategy.  
%\revision{Figure ?? shows the strategy employed by the expert in training.  The vertical gray lines represent when the environment changed; dots represent when a query was made.  The horizontal axis shows time in number of turns or iterations, and the vertical axis shows learning performance on the validation set after the completion of turn $t$.  As observed from the graph, using $\textbf{w}$, the imitation learner was able to closely match the expert's questioning strategy.}
%The expert provided three demonstrations of questioning sessions (learning episodes) in the training scenario.  
Environmental conditions were held constant across demonstrations and IRL training. Values changed for testing were time and budget allocation and frequency of environmental change.

%\kalesha{Provide additional context about IRL training process (see rebuttal to R2).}

% Learning Curves for Each Task
\begin{figure}[tb]
	\centering
	%\vspace{2mm}
	%\includegraphics[width=0.85\columnwidth]{figures/learning_curves_pack-lunchbox-task_exp1.png}
	\|*
	\begin{subfigure}[b]{0.85\columnwidth}
		\centering
		\includegraphics[width=\textwidth]{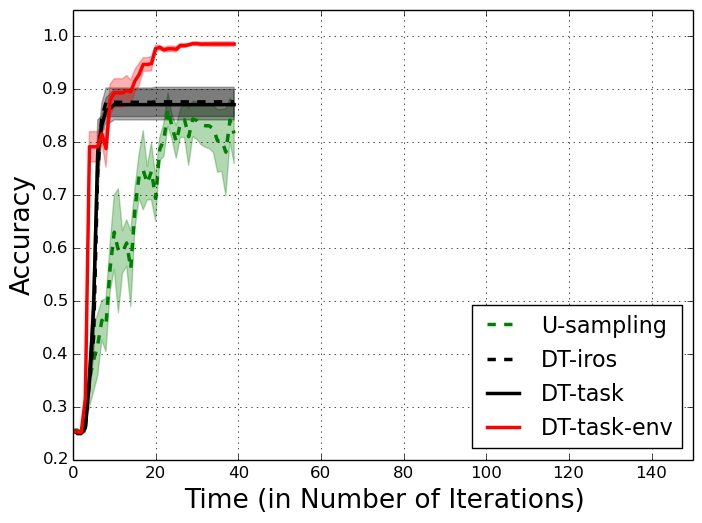} %task1_stat_sig_bar_chart_iter15
		\caption{Task 2: Constrained \textit{Time}}
		\label{fig:task2_exp3}
	\end{subfigure}\\
	*/
	\begin{subfigure}[b]{0.72\columnwidth}
		\centering
		\includegraphics[width=\textwidth]{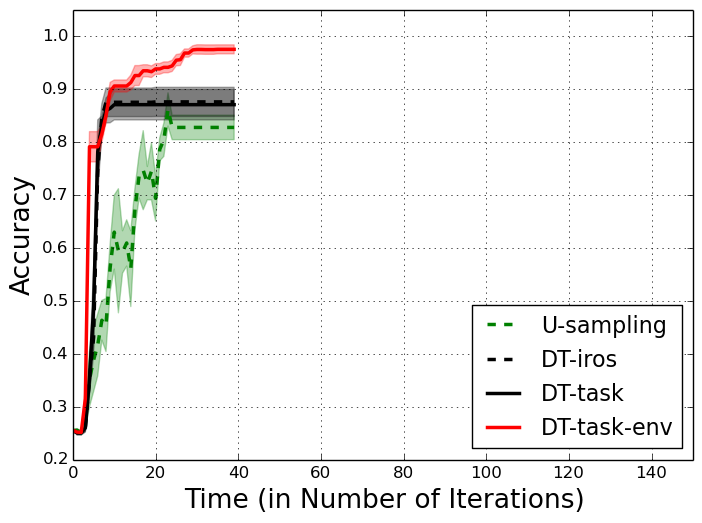}
		\caption{CONDITION 1: Constrained \textit{Time} \& \textit{Budget}}
		\label{fig:task2_exp1}
	\end{subfigure}
	\caption{Pack Lunchbox Task. Shows performance (test accuracy with standard error) on a separate task, under the most constrained experimental condition: \textit{both} time \textit{and} query budget constrained.}\label{fig:task2_learning_perf} 
	\vspace{-5mm}	 	 
\end{figure}

\subsection{Results}
%\revision{Still need to clarify environmentally constrained scenarios for testing (AL).  
%Towards testing our hypotheses, we sought out to investigate two types of generalization: (1) how the learned strategy generalizes to \textit{different} environmentally constrained conditions and (2) how the learned strategy generalizes to other tasks.  
%Continue here...}

To test our hypotheses, we first investigated generalization to \textit{different} environmentally constrained conditions. The goal was to vary time on one axis and resources (query budget) on the other axis.  In order to have a feasible stopping point, we could not truly allow unlimited time or query budget; however, as defined here, the \textit{constrained} versus \textit{unconstrained} parameters denote an order of magnitude difference in allocation.  All action types were assigned an a-priori cost, which is 2 for demo queries and feature subset queries, 1 for label queries, and 0 for no query.  This can be assigned in any way desired, but for our purposes, was intended to map roughly with the cognitive load required by the teacher in answering a particular type of question.  Figure \ref{fig:prepare_lunch_learning_curves} shows learning curves for each combination of time and resource parameters.  For each task, learning curves are averaged over 5 runs, each run sampling from a different pre-generated task training data sample.   Examining the subfigures: from left to right, time allocation is increased (from 40 to 150) and from bottom to top, query budget is increased (from 25 to 500). %Thus, in fig \ref{fig:prepare_lunch_experiment1}, both time and budget are constrained (budget 25, time 40).  This condition is the most similar to the training scenario (budget 15, time 30).  In fig \ref{fig:prepare_lunch_experiment2}, only query budget is constrained (budget 25, time 150).  In fig \ref{fig:prepare_lunch_experiment3}, only time is constrained (budget 500, time 40).  Fig \ref{fig:prepare_lunch_experiment2} represents the unconstrained condition (budget 500, time 150). 
Thus fig \ref{fig:prepare_lunch_experiment1} is the most constrained (budget 25, time 40) and most similar to the training scenario (budget 15, time 30).    
Overall, the left half roughly corresponds to the agent being given approximately 20 questions, assuming the most costly queries are minimized, whereas the right half corresponds to the agent being given unlimited queries during the time allocated. In testing, once a strategy has exceeded its query budget, it is no longer allowed to make queries, representing complete resource consumption.  Thus for the remainder of the episode, it must select the no-query action at no cost.  It should also be stated that we assume at most one query per turn.

% Learning Curves for Each Task
\begin{figure}[tb]
	\centering
	%\vspace{2mm}
	\begin{subfigure}[b]{0.7\columnwidth}
		\centering
		\includegraphics[width=\textwidth]{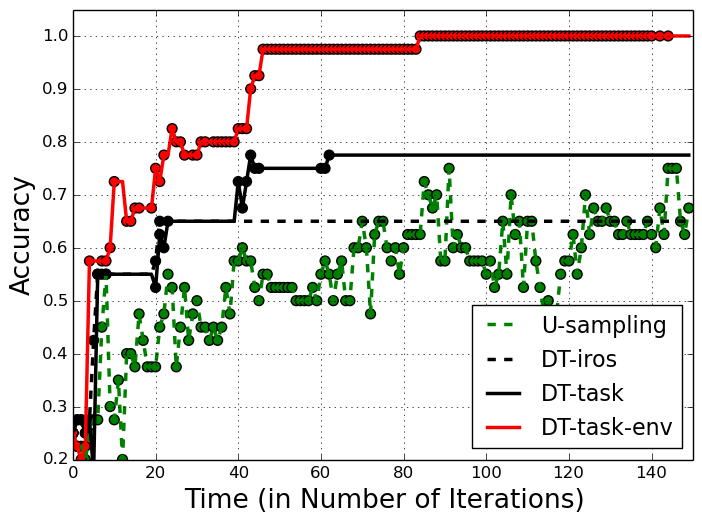}
		\caption{CONDITION 4: Unconstrained}
		\label{fig:task1_condition4_behavior}
	\end{subfigure}
	\begin{subfigure}[b]{0.7\columnwidth}
		\centering
		\includegraphics[width=\textwidth]{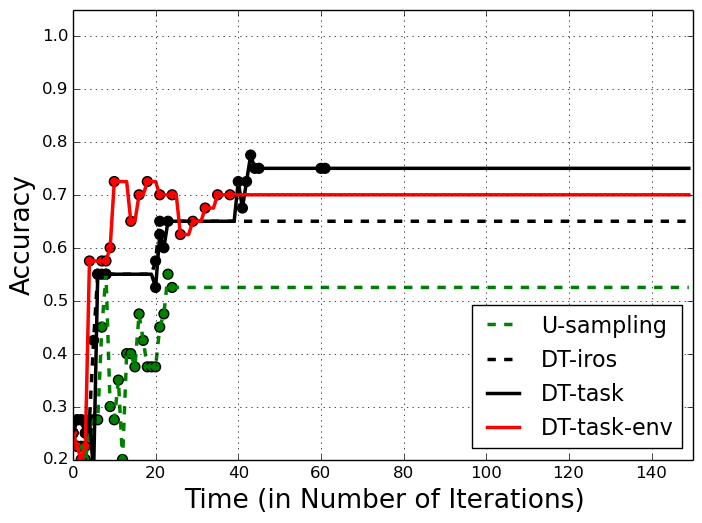}
		\caption{CONDITION 2: Constrained \textit{Budget}}
		\label{fig:task1_condition2_behavior}
	\end{subfigure}
	\caption{Questioning Behavior of each Strategy in Prepare-Lunch Task for \textit{one} training sample. Dots indicate when a query is made. }\label{fig:query_strategy_behavior} 
	\vspace{-3mm}	 	 
\end{figure}

%\kalesha{Discuss findings and how this relates to each hypothesis. And the statistical improvement in using DT-task-env (well says Mann-Whitney U Test).  Then move into understanding \textit{why} such domination occurs and explain Figure \ref{fig:query_strategy_behavior}. Then, mention the next task. Finally, wrap up with the overall insights extracted and implications of the work.}

We used the Mann-Whitney U-test to perform pairwise statistical comparisons of our experimental approach (DT-task-env) with each of the baseline approaches at the end of the allocated learning time, for all four experimental conditions.  A one-tailed test was conducted, as the goal was to understand if the experimental environmentally-aware approach \textit{improves} performance over the baseline task-centric approaches to AL.  In conditions 1, 3, and 4, DT-task-env statistically outperforms \textit{all} other approaches by the end of the learning time.  In condition 2, it statistically outperforms all approaches except DT-task, compared to which it performs equivalently.  Using the second (pack-lunchbox) task, we were able to replicate our results in another domain.  We only show the most constrained condition (limited time \textit{and} query budget) in Figure \ref{fig:task2_learning_perf}.  %, also averaged over five active learning runs for each strategy.  
Overall, there is a clear pattern.  DT-task-env \textit{always} performs \textit{at least} as well as the task-centric baselines but moreover \textit{dominates} task-centric approaches under most of the environmental conditions examined.  This confirms our second hypothesis that a questioning strategy learned through imitation of an expert in one environment \textit{can} be used to generalize to other constrained environments.  Our first hypothesis however is \textit{not} supported by findings from Condition 2.

To better understand what behavior leads to these findings, we analyze learning episodes from one training data sample under two different experimental conditions in Figure \ref{fig:query_strategy_behavior}. 
The dots represent points where queries were made for the given strategy.  We find that given a limited query budget and ample time (figure \ref{fig:task1_condition2_behavior}), DT-task and DT-task-env employ very similar \textit{conservative} strategies. Both use most of their budget closer to the beginning of the episode, as they attempt to build initial concept models, but also attempt to modulate budget consumption with rate of environmental change.  However, when the agent is allowed to ask unlimited queries given the same time frame (figure \ref{fig:task1_condition4_behavior}), these two strategies behave very differently.  Whereas the task-centric learners employ exactly the same strategy (because they have no ability to reason about environmental constraints), DT-task-env employs a very \textit{liberal} strategy.  In fact, it makes at least an order of magnitude more queries both than DT-task and DT-iros (134 versus 22 and 9), largely accounting for its complete domination over the other strategies.  Also notably, U-sampling asks a question at \textit{every} turn until it exceeds its budget, since it is able to modulate \textit{neither} for environmental change \textit{nor} for external constraints.  It also does not have the capability to autonomously select a feature subset query, so it must rely upon computational feature selection for solving its learning problem.  The decision-theoretic approaches, by contrast, are able to reason about and request an FSQ early in the episode, making them significantly more sample efficient.
%\revision{Continue here...}

The key implication of all of the experimental findings is that the DT-task-env strategy has the ability to effectively \textit{adapt} its questioning behavior \textit{both} to the rate of environmental change (like DT-iros and DT-task) \textit{and} to time and resource constraints imposed externally.  In more realistic environments, this is compelling as it gives human partners the capability to specify their own time and cognitive load constraints, with the understanding that the agent can integrate this knowledge into its reasoning about the learning task.

\section{Discussion}
%\kalesha{Add discussion about (1) limitations of data collection and (2) providing more insights about IRL training process (see rebuttal to R3).}

This work contributes a new cognitive capability for active learners in more realistic contexts, by enabling them with a policy to trade off learning objectives with environmental constraints.  Yet, it is not without its limitations.  The properties of environmental change captured by the object datasets used, serve only as a proxy for partial observability encountered in the real world.  In realistic environments, scenes are often cluttered and scene change happens continuously.  

Additionally, the active learning models contributed are intended for use by any type of artificial agent and thus agnostic to an agent's embodiment.  However, \textit{what} an agent decides to ask cannot always be decoupled from \textit{how} it must use its embodiment to execute the query.  For example, query cost may increase if a query requires fine-grained manipulation that is difficult for the agent to maneuver or simply takes a longer time to generate than other queries.  Ideally, this should be incorporated into the agent's decision function.  

Future work could explore how to adapt these methods, given more complex perceptual data or agent embodiment as input for the questioning framework.

%Regarding the Inverse Reinforcement Learning training process... \revision{not really sure how to pull this in?}

%This more complex perceptual data states is not considered, as the focus of the work is on the reasoning problem for the learner, but is import.

\|*
\subsection{Still to do}
\revision{
additional major points to discuss...
\begin{itemize}
	\|*
	\item hypotheses being tested:
	\begin{enumerate}
		\item Reasoning \textit{additionally} about environmental context can enable an active learning agent to adapt its query selection strategy and improve its learning performance under constrained conditions.
		%\item Reasoning \textit{additionally} about environmental context can enable an active learning agent to adapt its query selection strategy, in accordance with time and resources allocated to it, and improve its learning performance under constrained conditions.
		\item Imitation Learning can be used to infer an expert's strategy for managing time and resources allocated for solving a given learning problem, then generalized to other constrained environments. \kalesha{ask yourself: what graphs/visualizations would be needed to support this claim and can i reasonably produce them in the time left before IJCAI?}
		\begin{enumerate}
			\item show graph with queries visualized as dots which depict both expert's strategy and learned policy on the training scenario. then show graph with dots for learned policy under each of the different constrained conditions.  speaks to agent's ability to generalize.
			\item show agent's ability to generalize strategy under different constrained conditions
		\end{enumerate}
		%\item Imitation Learning can be used to infer how an expert trades off learning goals with environmental constraints and thus manages time and resources allocated for solving the given learning problem. \kalesha{ask yourself: what kinda graphs/visualizations would this entail?}
		%\item Reasoning about environmental context (in addition to learning objectives) can improve an active learning agent's ability to learn, given the agent exists within a dynamic environment with time or resource constraints being imposed on it.
	\end{enumerate} 
	\item evaluation metrics for assessing competing learning agents (recognition accuracy on validation set).
	*/
	\item algorithms being examined/explored. clearly distinguish baseline approaches from experimental approach(es). 
	\begin{enumerate}
		\item BASELINE 1: standard AL baseline - uncertainty sampling
		\item BASELINE 2: decision-theoretic arbitrator (task-centric, manually tuned, published in IROS)
		\item BASELINE 3: decision-theoretic arbitrator (task-centric, manually tuned, updated version w/new decision feature)
		\item EXPERIMENTAL: decision-theoretic arbitrator (task AND environment, learned through IRL)
	\end{enumerate}
	\item experimental design, learning task used
	\begin{enumerate}
		\item \kalesha{Should clarify experimental design for BOTH training (IRL) and testing (AL)}
		\item During training of environmentally sensitive approach, the maximum number of iterations was set at 100, and we selected the set of weights $\textbf{w}^*$ that performed best on the validation set.
	\end{enumerate}
\end{itemize}
}
*/

%\kalesha{Discussion seems in order now, though maybe no room for an additional section header.}

%\vspace{-2mm}
\section{Conclusion}
%\revision{Write a few concluding remarks here.}

This work contributed a first exploration and novel computational approach for solving the problem of active learning under externally imposed time and resource constraints.  %It also contributed a novel approach for trading off learning objectives with environmental constraints, 
Imitation of an expert questioner's policy was used to learn to weight the diverse set of decision criteria for an environmentally-aware active learner.  Experiments were conducted under various environmentally constrained conditions and on two concept learning tasks.  Key findings show the experimental approach presented statistically outperformed a standard uncertainty sampling baseline and the strictly task-centric active learners under most environmental conditions; thus representing a promising alternative for active learners in more realistic human environments.

%\FloatBarrier
%\FloatBarrier
%\FloatBarrier
%\pagebreak
\bibliographystyle{named}
\bibliography{references,sonia}

\end{document}